\documentclass[runningheads]{llncs}
\usepackage[T1]{fontenc}
\usepackage{graphicx}
\usepackage{amsmath}
\usepackage{booktabs}
\usepackage{enumitem}
\usepackage{sectsty}
\usepackage{siunitx}
\usepackage[table]{xcolor}
\newif\ifshowcomments
\showcommentstrue 

\begin{document}
\title{What Makes a Dribble Successful?\\Insights From 3D Pose Tracking Data}
%
%
\author{Michiel Schepers\inst{1} \and
Pieter Robberechts\inst{1,2}\orcidID{0000-0002-3734-0047} \and
Jan Van Haaren\inst{1,3}\orcidID{0000-0001-7737-5490} \and
Jesse Davis\inst{1,2}\orcidID{0000-0002-3748-9263}}
\authorrunning{M. Schepers et al.}
%
\institute{KU Leuven, Dept. of Computer Science, B-3000 Leuven, Belgium
\email{michiel.schepers@student.kuleuven.be, \{pieter.robberechts,jesse.davis\}@kuleuven.be}
\and
Leuven.AI - KU Leuven Institute for AI, B-3000 Leuven, Belgium
 \and
Club Brugge KV, B-8300 Knokke-Heist, Belgium\\
\email{jan.vanhaaren@clubbrugge.be}}
\maketitle              
\begin{abstract}
Data analysis plays an increasingly important role in soccer, offering new ways to evaluate individual and team performance. One specific application is the evaluation of dribbles: one-on-one situations where an attacker attempts to bypass a defender with the ball. 
While previous research has primarily relied on 2D positional tracking data, this fails to capture aspects like balance, orientation, and ball control, limiting the depth of current insights.
This study explores how pose tracking data---capturing players' posture and movement in three dimensions---can improve our understanding of dribbling skills. 
We extract novel pose-based features from 1,736 dribbles in the 2022/23 Champions League season and evaluate their impact on dribble success.
Our results indicate that features capturing the attacker's balance and the alignment of the orientation between the attacker and defender are informative for predicting dribble success. Incorporating these pose-based features on top of features derived from traditional 2D positional data leads to a measurable improvement in model performance.
\end{abstract}
\section{Introduction}\label{sec:introduction}

Although soccer is a team sport, matches are often decided by individual actions, such as one-on-one dribbles between an attacker and a defender. Therefore, there is a continuous search for new ways to analyze, evaluate, and improve the technical skills involved in these direct confrontations. 

Currently, analytical models that assess individual actions like one-on-one dribbles rely predominantly on 2D tracking data~\cite{meerhoff2019exploring,brink2023measuring}, recording player and ball positions on the pitch’s horizontal plane over time. While these models have yielded valuable insights, they fail to capture essential factors, such as balance, posture, and ball control, all of which influence the success of a dribble. Moreover, existing models often overlook the subtleties of the interaction between attacker and defender, limiting their predictive accuracy and interpretability. For instance, an attacker may attempt to pass a defender on the side opposite their supporting leg to exploit a momentary imbalance.

With the continuous evolution of tracking technology, clubs increasingly have access to 3D pose tracking data, which captures the coordinates of key anatomical landmarks on players’ bodies in real time~\cite{hawkeye2024}. While currently the main use of this data is in officiating, particularly for accurately determining offside decisions~\cite{fifa2023semiautomated}, its potential extends far beyond this. 

In this paper, we propose a novel approach to analyzing dribbling success by extracting meaningful 3D features from pose tracking data. These features capture the player's ball control, balance and posture, and spatial interactions with the defender. We evaluate their relevance in explaining dribble success and examine whether incorporating these 3D biomechanical features into a logistic regression model improves predictive accuracy compared to models based solely on traditional 2D data.

Using a comprehensive dataset of 1,736 one-on-one dribbles from the 2022/23 UEFA Champions League season, our results demonstrate that integrating 3D features leads to a clear improvement in predictive performance. SHAP value analysis highlights that defender posture and attacker balance are among the most impactful variables, underscoring the added value of 3D biomechanical information. These findings suggest that 3D pose tracking data can significantly enhance the analysis of individual player actions, offering new perspectives for performance evaluation, scouting, and player development. 
\section{Background and Related Work}

\paragraph{Pose Tracking Data}

\begin{figure}[h!]
\centering
\includegraphics[width=0.25\textwidth]{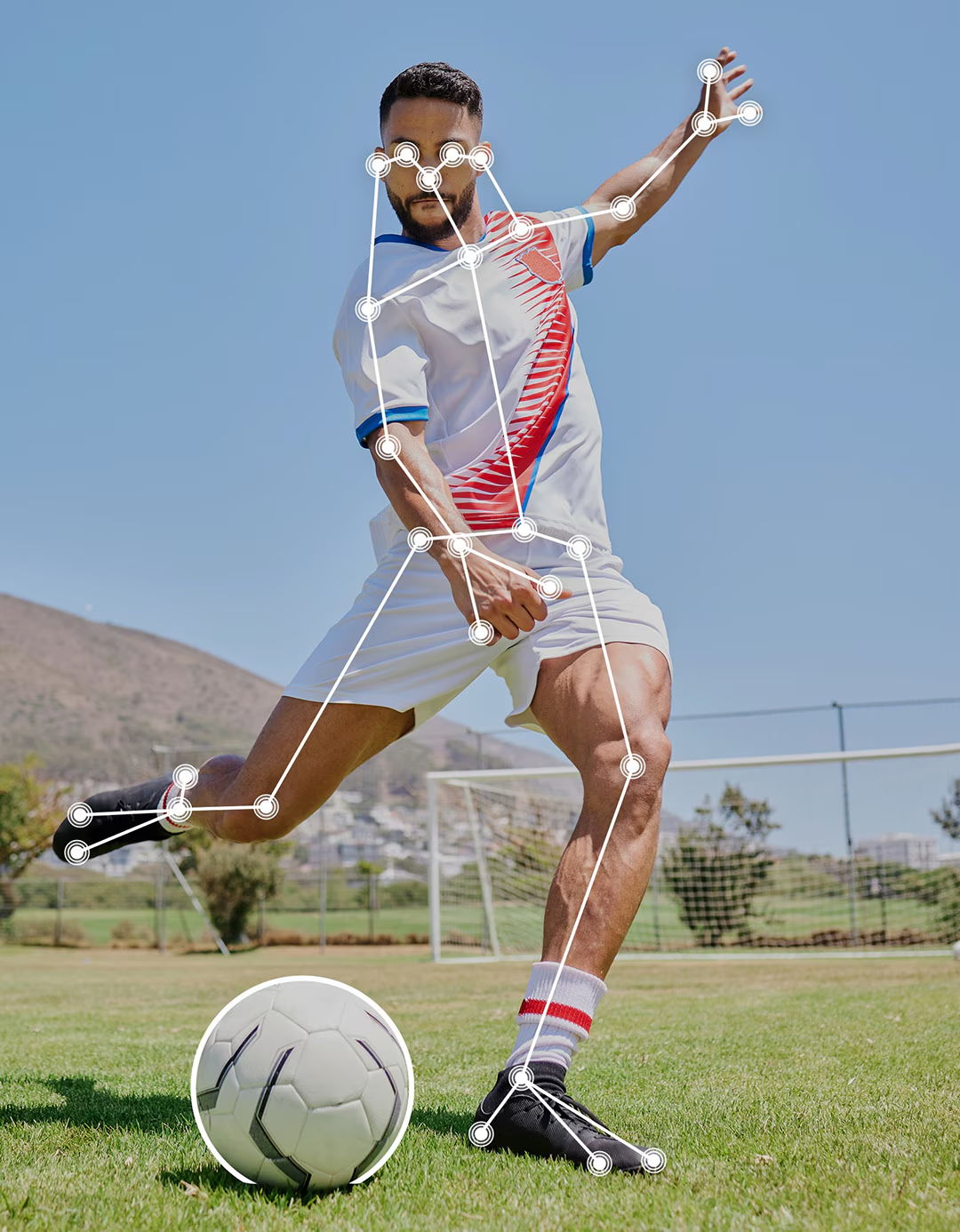}
\caption{The 29 anatomical points whose positions are tracked by the SkeleTRACK system (from \cite{hawkeye2024}).}
\label{fig:skeletrack}
\end{figure}

Until recently, soccer data primarily consisted of 2D information~\cite{rahimian2022:opticaltracking}, tracking only the {XY} positions of players and the ball on the field, which limited analysis to a flat perspective. However, recent technological advances have transformed this approach with the introduction of 3D pose (or, skeleton) tracking data. This technology captures not only players’ positions but also their body posture and movements in three dimensions, using a network of ultra-high-speed cameras. These recordings are analyzed  with AI algorithms to reconstruct the 3D pose of each player~\cite{arbues2021}. This is done by calculating the coordinates of a number of anatomical landmarks on the player’s body, such as the shoulders, hips, ankles, and knees. Figure~\ref{fig:skeletrack} provides a visual representation of what is tracked by such a system. 

\paragraph{Analysis and Modeling of Dribbles}
Various studies have aimed to model dribbling dynamics and quantify player skills based on 2D data. A notable contribution is the VAEP model introduced by Decroos et al.~\cite{decroos2019actions}, which estimates the value of individual player actions, including dribbles. The model predicts the probability that a team will score or concede a goal within a few subsequent actions, defining the value of a dribble as the change in expected scoring chances before and after the action. While this approach provides an objective measure of a dribble’s impact, it offers limited insight into the underlying technical skills.

Complementing this, Brink et al.~\cite{brink2023measuring} focus specifically on the dynamics of one-on-one dribbles, introducing a behaviorally inspired model that simulates attacker-defender interactions by assigning physical parameters (such as speed and acceleration) and behavioral traits (such as attacker aggressiveness and defender hesitation) to players. Though valuable for exploring  the conditions under which dribbles succeed or fail, this approach remains limited by 2D data, which omit key biomechanical factors such as posture and balance.

Beyond data-driven approaches, sports science research has long investigated determinants of dribbling success through controlled experiments. For instance, Zago et al.~\cite{zago2016dribbling} conducted tests with youth players measuring biomechanical variables like center of mass and step length, capturing physical dynamics not accessible via positional tracking data.

This paper aims to bridge these two research domains by extracting bio\-mechanically-informed
 features from 3D pose tracking data and integrating them into quantitative models.


\section{Data, Preprocessing and Feature Engineering}\label{sec:data}

In this section, we first describe our data set and how we selected the dribbles that will be analyzed.

\subsection{Data Set}

For this study, we have used a dataset  of 125 games from the 2022/23 UEFA Champions League season,  provided by Club Brugge KV. For each game, the data consists of 2D positional tracking data, 3D pose tracking data, and event data. The positional and pose tracking data were collected by Hawk-Eye using the SkeleTRACK system and record the location of the ball, the centroid of each player and 29 anatomical landmarks for each player (Figure~\ref{fig:skeletrack}), sampled at 25Hz. The event data was collected by Stats Perform and contains a log of all on-the-ball actions that occurred during the game. Before analysis, the raw data must be prepared through synchronization and filtering to isolate relevant one-on-one dribbles. The pre-processing steps are detailed below, resulting in a dataset of 1,736 dribbles.


\subsection{Data Preprocessing}

First, event data is integrated with the 2D and 3D tracking data by linking the start and end timestamps of each event to the corresponding frames. The start timestamp corresponds to the moment the previous event ends, and the end timestamp is the start of the next event. We do not observe major misalignments between both data providers, making complex synchronization procedures~\cite{van2023etsy} unnecessary. 


Second, missing pose data (typically caused by player occlusion or brief exits from the camera frame) is addressed using cubic spline interpolation~\cite{chapra2011numerical} with natural boundary conditions. This yields smooth, realistic motion trajectories, enabling robust feature extraction.

\subsection{Dribble Filtering}
The construction of a one-on-one dribble dataset starts from all \texttt{TAKE\_ON} events in the event data, which indicate {\it ``an attempt by one player to dribble past an opponent''}. Because not all such events represent meaningful one-on-one dribbles, we apply additional filtering criteria:

{
\sisetup{
  round-mode = none
}
\begin{enumerate}
    \item The player must cover at least \SI{2}{\metre} and the dribbles must last at least \SI{2}{\second}. This requirement ensures sufficient movement to extract informative features.
    \item The average ball speed must exceed \SI{1.5}{\metre\per\second} during the dribble. This threshold filters out slow actions or duels lacking real progression, focusing only on genuine one-on-one dribbles where the attacker actively attempts to advance.
    \item The maximum ball speed must remain below \SI{11}{\metre\per\second}. This criterion helps eliminate incorrectly labeled events: ball speeds above this threshold typically indicate a pass or shot~\cite{zonalsports_fastest2025}.
    \item The endpoint must be closer to the opponent’s goal than the start, ensuring attacking intent. Backward dribbles typically serve different purposes (e.g., ball retention or repositioning) and are therefore less relevant in the context of one-on-one offensive analyses.
\end{enumerate}
}

These rules were developed through an iterative process involving manual video inspection to validate coverage and quality. Applying them yields a dataset of 1,736 dribbles: 690 successful and 1,046 unsuccessful, with success defined as the attacker retaining possession after beating the defender~\cite{statsperform_takeon}. Each dribble is also linked to its primary defender, defined as the nearest opponent at the end of the dribble.
\subsection{Feature Engineering for Dribble Analysis}\label{sec:features}

To characterize each dribble, we extracted 8 features from the 2D positional tracking data and 6 features from the 3D pose data (Table~\ref{tab:features}). 

The 2D features~\cite{brink2023measuring,oonk2024dribble} capture the dribble's location, the movement of the players, and their interactions, but in a relatively limited way. To describe the dribbles's position on the field, we included the distance to the nearest sideline and to the opponent’s goal. Attacker and defender movement is captured by maximum speed, acceleration, and directional changes~\cite{kai2021new}. Interaction is quantified through relative speed (attacker minus defender) and a pressure score based on whether the attacker entered the defender’s pressure zone, defined following Andrienko et al.~\cite{andrienko2017visual}.

Our novel 3D features capture the player's ball control, balance, and more detailed player interactions. To characterize ball control, we computed the average distance between the ball and the attacker's nearest foot, normalized by defensive pressure. By normalizing ball control in this way, the context of the dribble is taken into account, providing a more realistic measure of the player’s skill depending on the situation in which they find themselves. We also calculated the frequency of ball touches, defined as local minima in the foot–ball distance trajectory. A higher frequency of ball contact generally indicates a more technical dribbling style, while a lower frequency suggests players who rely more on speed to get past their opponent. To assess the attacker's balance, we first estimated the player's center of mass (CoM) using an anthropometric model~\cite{winter2009biomechanics}. Based on the CoM, we then describe the player's balance with two features: imbalance (computed as the horizontal offset between the CoM and the midpoint between their feet) and torso lean angle (computed as the angle between the torso vector from hip to neck and the vertical axis). Finally, to capture the interaction between the attacker and defender, we consider whether the attacker attempts to pass on the side opposite the defender’s weighted leg, as well as the alignment of each player’s body direction relative to the vector connecting them.

Additional features and aggregations were considered but were not retained to avoid multicollinearity. For example, maximum speed and acceleration of attacker and defender were preferred over their averages because dribbles are explosive actions best characterized by peak values. Similarly, average pressure from nearby opponents was kept as it provides a more representative overall context than peak pressure. Appendix~\ref{app:features} provides a detailed description of all considered features.

\begin{table}
\centering
\caption{Extracted features, categories, and descriptions.}
\label{tab:features}
\small
\begin{tabular}{@{}p{4cm}p{3cm}p{6.5cm}@{}}
\toprule
\textbf{Feature} & \textbf{Category} & \textbf{Description} \\ 
\midrule
Dist. to sideline (Atk.) & 2D Spatial & Distance from attacker’s start position to nearest sideline (\si{m}) \\
Dist. to goal (Atk.) & 2D Spatial & Distance from attacker’s start position to opponent’s goal (\si{m}) \\\midrule
Max speed (Atk.) & 2D Kinematics & Peak speed reached by the attacker (\si{m/s}) \\
Max speed (Def.) & 2D Kinematics & Peak speed reached by the defender (\si{m/s}) \\
Max acceleration (Atk.) & 2D Kinematics & Highest acceleration of the attacker (\si{m/s^2}) \\
Max direction change (Atk.) & 2D Kinematics & Maximum angular change in attacker’s movement direction (\si{\degree})\\\midrule
Max relative speed & 2D Interaction & Maximum difference in speed between attacker and defender (\si{m/s})\\
Avg. pressure score & 2D Interaction & Average pressure exerted by nearby defenders on attacker (unitless) \\\midrule
Avg. norm. ball-foot dist. & 3D Ball Control & Average distance between ball and attacker’s nearest foot, normalized by defensive pressure (unitless)\\
Ball touch frequency & 3D Ball Control & Number of ball touches per second, detected as local minima in foot-ball distance (\si{\text{touches}\per\second})\\\midrule
Avg. imbalance & 3D Pose & Average distance in the XY plane between the attacker's center of mass and the midpoint between the two feet (unitless)\\
P90 torso lean angle & 3D Pose & 90th percentile of attacker's lean angle relative to vertical, based on a vector from the neck to the hip midpoint. (\si{\degree})\\\midrule
Pass side vs. weighted leg & 3D Interaction & Whether the attacker attempts to pass the defender on the side opposite the defender's weighted leg, determined by comparing center of mass distance to each heel (binary)\\
Defender stance angle & 3D Interaction & Alignment of each player's body direction relative to the vector connecting them (\si{\degree})\\
\bottomrule
\end{tabular}
\end{table}

\section{Empirical Analysis of Dribble Success}
Our empirical analysis of the features presented in Section~\ref{sec:features} aims to answer the following two research questions:

\begin{enumerate}[label=\textbf{Q\arabic*:}, leftmargin=*, align=left, itemsep=1em]
	\item \textbf{Can the addition of 3D pose-based features improve the prediction accuracy of existing 2D dribble outcome models?}
	\item \textbf{Which features extracted from 3D pose tracking data are relevant for describing dribbles, and how do they impact the success rate of those dribbles?}
\end{enumerate}

The first question focuses on examining the practical value of pose-based descriptors of dribbles in improving predictive models,  while the second aims to understand how these features affect the success of dribbles. The next sections discuss the methodology and present the results.

\subsection{Methodology}
We compare a simple location-based baseline with a logistic regression model trained on both 2D tracking and 3D pose-derived features.

\paragraph{Baseline Model}
To capture the influence of a dribble's location\footnote{It is generally more difficult to successfully complete a dribble compared to areas further from goal}, the pitch is divided into an $8\times5$ grid. For each cell, we compute the success rate as the proportion of successful dribbles starting in that zone. Predictions are made by assigning a new dribble the success rate of its corresponding cell. This naive model ignores contextual or biomechanical factors.

\paragraph{Predictive Model}
The main model uses logistic regression~\cite{sklearn} to predict dribble outcomes based on both 2D and 3D features. Logistic regression was chosen for its interpretability and its ability to reveal the individual contribution of each feature. All features were standardized by removing the mean and scaling to unit variance to ensure equal treatment. The model was trained with L2 regularization using the \texttt{SAGA}~\cite{defazio2014:saga} solver, with a maximum of 1000 iterations. To address class imbalance, sample weights were set inversely proportional to class frequencies.


\paragraph{Evaluation}
We assess model performance using five-fold cross-validation in terms of precision, recall, F1-score, and Brier score.

\subsection{Q1: Impact of 3D Features on Model Performance} \label{sec:results_model}

To assess the contribution of different feature sets to model performance, we conduct an ablation study across three configurations: a baseline model using only the dribble location, a 2D model incorporating spatiotemporal tracking features, and a full model combining both 2D and 3D pose-based features. Table~\ref{tab:model_comparison} reports the results as average scores across folds, together with the standard deviations.

The baseline model achieves the highest precision but suffers from low recall, reflecting its limited ability to detect successful dribbles in an imbalanced dataset. Introducing 2D features significantly improves recall and F1 score, highlighting the added value of basic kinematic features and features that capture the level of defensive pressure. Adding 3D pose features yields further gains across all metrics. Although the performance increase is modest, it indicates that pose-based information captures additional aspects of dribbling skill that are not represented in 2D tracking data alone.

\begin{table}[ht]
	\centering
	\caption{Model performance on dribble outcome prediction, demonstrating that adding 3D pose features improves predictive performance.}
	\begin{tabular}{l
			@{\hskip 12pt} S[table-format=1.2(2)]
			@{\hskip 12pt} S[table-format=1.2(2)]
			@{\hskip 12pt} S[table-format=1.2(2)]
			@{\hskip 12pt} S[table-format=1.4(4)]}
		\toprule
		\textbf{Model}   & \textbf{Precision $\uparrow$} & \textbf{Recall $\uparrow$} & \textbf{F1 Score $\uparrow$} & \textbf{Brier score $\downarrow$} \\
		\midrule
		Baseline         & \bfseries 0.55(8)             & 0.26(4)                    & 0.35(4)                      & \bfseries 0.2356(78)              \\
		2D Features Only & 0.46(2)                       & 0.53(2)                    & 0.49(2)                      & 0.2444(55)                        \\
		2D + 3D Features & 0.47(2)                       & \bfseries 0.57(4)          & \bfseries 0.52(3)            & 0.2429(28)                        \\
		\bottomrule
	\end{tabular}
	\label{tab:model_comparison}
\end{table}

\subsection{Q2: Key Factors Influencing Dribble Success} \label{sec:results_shap}
\begin{figure}
	\centering
	\includegraphics[width=.8\textwidth]{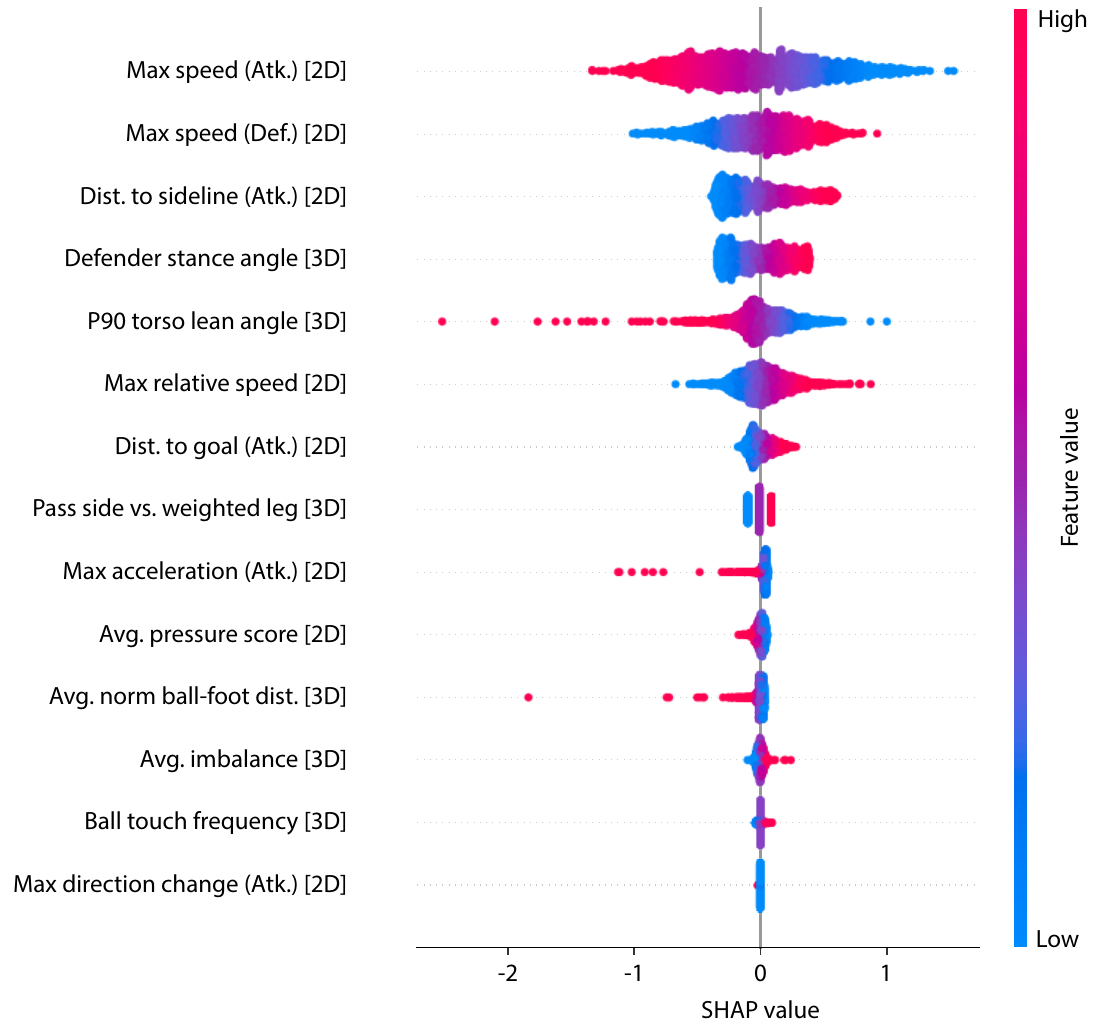}
	\caption{SHAP values for the 2D and 3D features used for predicting dribble success. Positive SHAP values (right) indicate features that increase the probability of a successful dribble, while negative values (left) decrease it.} \label{fig:shap}
\end{figure}

To better understand the contribution of individual features to the model's predictions, SHAP values were computed across all features. Figure~\ref{fig:shap} presents the SHAP values for the 14 features used for predicting dribble success. The sign and magnitude of these values indicate the direction and strength of each feature's effect on the predicted probability of a successful dribble. Specifically, features with higher SHAP values concentrated on the right suggest that increased feature values enhance the likelihood of a successful dribble, whereas those skewed to the left indicate the opposite.

Among all features, maximum speed emerges as a key determinant, both for the attacker and defender. This finding is consistent with prior models such as Brink et al.~\cite{brink2023measuring}, which emphasize speed and acceleration as fundamental predictors. However, somewhat counterintuitively, a higher maximum speed corresponds with a reduced probability of a successful dribble. This suggests that while speed is important, other factors such as ball control may moderate its positive effect, aligning with previous observations from isolated experiments.

Additionally, spatial context matters: dribbles performed further from the sideline have a higher likelihood of success, underscoring the importance of positional factors in outcome prediction.

In contrast, features capturing attacker-defender interaction appear less influential overall. This finding challenges the assertion by Oonk et al.~\cite{oonk2024dribble} that only interactive features significantly affect dribble outcomes. Nevertheless, the two interactive features identified highlight critical aspects of attacker-defender interaction. Notably, elevated average defensive pressure during the dribble exerts a negative influence on success probability. Conversely, moments when the attacker exhibits a substantial relative speed advantage over the defender coincide with increased chances of dribble success, corroborating findings from Oonk et al.~\cite{oonk2024dribble}.

Among the most impactful 3D features, the defender's posture is paramount. A larger angle between the torso vectors of attacker and defender (Body Direction Alignment) signals a greater chance of a successful dribble, highlighting the role of defensive orientation. Similarly, the torso lean angle significantly affects model predictions; large lean angles measured during the dribble are associated with a higher likelihood of failure. These insights emphasize the importance of balance and body positioning. Finally, average normalized ball control, ball touch frequency and the imbalance metric appear to have a minimal effect on the model’s predictions.

\section{Conclusions}
This paper aimed to analyze dribbling in professional football using 3D pose tracking data from Hawk-Eye, focusing on identifying relevant 3D features that describe dribbles and their impact on success. A methodology was developed to extract meaningful 3D features related to ball control, balance and attacker-defender interactions. The analysis showed that balance metrics and the alignment with the defender's posture strongly correlate with dribble success, while ball control had limited impact. Incorporating these 3D features into a logistic regression model alongside classic 2D features such as player speed, field position, and defensive pressure yielded a modest but clear improvement in predictive performance. These findings demonstrate that 3D pose data provides valuable new insights for understanding and predicting dribble outcomes in football.
%
%
%
\bibliographystyle{splncs04}
\bibliography{mybibliography}

\begin{thebibliography}{10}
\providecommand{\url}[1]{\texttt{#1}}
\providecommand{\urlprefix}{URL }
\providecommand{\doi}[1]{https://doi.org/#1}

\bibitem{andrienko2017visual}
Andrienko, G., Andrienko, N., Budziak, G., Dykes, J., Fuchs, G.,
  Von~Landesberger, T., Weber, H.: Visual analysis of pressure in football.
  Data Mining and Knowledge Discovery  \textbf{31},  1793--1839 (2017)

\bibitem{arbues2021}
Arbués~Sangüesa, A.: A Journey of Computer Vision in Sports: From Tracking to
  Orientation-Based Metrics. Ph.D. thesis, Universitat Pompeu Fabra, Barcelona,
  Spain (2021),
  \url{https://repositori.upf.edu/items/835e902c-e861-4dcb-a02f-4fbcc332bc4e},
  phD thesis

\bibitem{zonalsports_fastest2025}
Benjamin, I.: Top 10 fastest players in the world [2025].
  \url{https://www.zonalsports.com/ranking/fastest-football-players} (2025),
  accessed: June 3, 2025

\bibitem{brink2023measuring}
Brink, L., Ha, S.K., Snowdon, J., Vidal-Codina, F., Rauch, B., Wang, F., Wu,
  D., L{\'o}pez-Felip, M.A., Clanet, C., Hosoi, A.E.: Measuring skill via
  player dynamics in football dribbling. Scientific Reports  \textbf{13}(1),
  19004 (2023)

\bibitem{chapra2011numerical}
Chapra, S.C., Canale, R.P., et~al.: Numerical methods for engineers, vol.~1221.
  Mcgraw-hill New York (2011)

\bibitem{decroos2019actions}
Decroos, T., Bransen, L., Van~Haaren, J., Davis, J.: Actions speak louder than
  goals: Valuing player actions in soccer. In: Proceedings of the 25th ACM
  SIGKDD international conference on knowledge discovery \& data mining. pp.
  1851--1861 (2019)

\bibitem{defazio2014:saga}
Defazio, A., Bach, F., Lacoste-Julien, S.: Saga: A fast incremental gradient
  method with support for non-strongly convex composite objectives. Advances in
  neural information processing systems  \textbf{27} (2014)

\bibitem{evangelos2012proprioception}
Evangelos, B., Georgios, K., Konstantinos, A., Gissis, I., Papadopoulos, C.,
  Aristomenis, S.: Proprioception and balance training can improve amateur
  soccer players' technical skills. Journal of Physical Education and Sport
  \textbf{12}(1), ~81 (2012)

\bibitem{fifa2023semiautomated}
FIFA: Semi-automated offside technology.
  \url{https://inside.fifa.com/innovation/world-cup-2022/semi-automated-offside-technology}
  (2023), accessed: April 16, 2025

\bibitem{kai2021new}
Kai, T., Hirai, S., Anbe, Y., Takai, Y.: A new approach to quantify angles and
  time of changes-of-direction during soccer matches. Plos one  \textbf{16}(5),
   e0251292 (2021)

\bibitem{meerhoff2019exploring}
Meerhoff, L.A., Goes, F.R., De~Leeuw, A.W., Knobbe, A.: Exploring successful
  team tactics in soccer tracking data. In: Joint european conference on
  machine learning and knowledge discovery in databases. pp. 235--246. Springer
  (2019)

\bibitem{oonk2024dribble}
Oonk, A., Buurke, T., Kempe, M., Lemmink, K.A.: Dribble like robben: What
  determines successfulness in 1-vs-1 actions in elite soccer? In: ECSS 2024
  (2024)

\bibitem{sklearn}
Pedregosa, F., Varoquaux, G., Gramfort, A., Michel, V., Thirion, B., Grisel,
  O., Blondel, M., Prettenhofer, P., Weiss, R., Dubourg, V., Vanderplas, J.,
  Passos, A., Cournapeau, D., Brucher, M., Perrot, M., Duchesnay, E.:
  Scikit-learn: Machine learning in {P}ython. Journal of Machine Learning
  Research  \textbf{12},  2825--2830 (2011)

\bibitem{rahimian2022:opticaltracking}
Rahimian, P., Toka, L.: Optical tracking in team sports: A survey on player and
  ball tracking methods in soccer and other team sports. Journal of
  Quantitative Analysis in Sports  \textbf{18}(1),  35--57 (2022)

\bibitem{hawkeye2024}
{Sony Group Corporation}: Hawk-eye: The evolution of sports officiating and
  analysis.
  \url{https://www.sony.com/en/SonyInfo/technology/stories/entries/20240411/hawkeye/}
  (Accessed: March 1, 2025)

\bibitem{statsperform_takeon}
{Stats Perform}: Opta event definitions.
  \url{https://www.statsperform.com/opta-event-definitions/} (2024), accessed:
  May 13, 2025

\bibitem{van2023etsy}
Van~Roy, M., Cascioli, L., Davis, J.: Etsy: A rule-based approach to event and
  tracking data synchronization. In: International Workshop on Machine Learning
  and Data Mining for Sports Analytics. pp. 11--23. Springer (2023)

\bibitem{winter2009biomechanics}
Winter, D.A.: Biomechanics and motor control of human movement. John wiley \&
  sons (2009)

\bibitem{yusuf2022effect}
Yusuf, M.Z., Rumini, R., Setyawati, H.: The effect of agility and balance
  training on dribbling speed in soccer games. Journal of Physical Education
  and Sports  \textbf{11}(1),  125--133 (2022)

\bibitem{zago2016dribbling}
Zago, M., Piovan, A.G., Annoni, I., Ciprandi, D., Iaia, F.M., Sforza, C.:
  Dribbling determinants in sub-elite youth soccer players. Journal of sports
  sciences  \textbf{34}(5),  411--419 (2016)

\end{thebibliography}
\appendix 
\paragraphfont{\bfseries}
\section{Feature Definitions and Formulae}
\label{app:features}

This appendix lists every feature used in our analysis, together with a brief explanation and the mathematical expression employed in the implementation. The following notations are used:
\begin{itemize}
\item $T$ — dribble duration; $N$ — number of frames.
\item $\mathbf{j}_{\cdot}$ — 3D joint coordinates returned by pose tracking.
\item Subscripts “att”, “def”, “bf” denote attacker, defender, ball–foot.
\item All spatial norms $\|\cdot\|$ are Euclidean distances in metres.
\end{itemize}
All integrals are evaluated numerically from frame‐wise samples, i.e.\ $\int_0^{T}x(t)\,dt\approx\frac{1}{N}\sum_{k=1}^{N}x_k\;\Delta t$.

\subsection{2D Spatial Features}
These features describe the location of the dribble on the pitch, which impacts how aggressively the defense is played. For dribbles occurring far from the goal or close to the sideline, a defender is less compelled to defend aggressively than for dribbles close to the goal.

\paragraph{Distance to sideline}
\[
d_{\text{side}} = \min\bigl( |y(t) - y_{\min}|,\; |y(t) - y_{\max}| \bigr)
\]
where $y(t)$ is the lateral position of the attacker at time $t$, and $y_{\min}$, $y_{\max}$ are the $y$-coordinates of the two sidelines.

\paragraph{Distance to goal}
\[
d_{\text{goal}} = \left\| \mathbf{p}(t) - \mathbf{g} \right\|,
\]
where $\mathbf{p}(t) = (x(t), y(t))$ is the attacker’s position at time $t$, and $\mathbf{g} = (x_{\text{goal}}, y_{\text{goal}})$ is the fixed 2-D coordinate of the center of the opponent’s goal.

\subsection{2D Kinematic Features}

These features describe the speed and acceleration of a player during the dribble. They are computed both for the attacker and defender, independently of each other.

\paragraph{Average speed}
\[
\bar{v} = \frac{1}{T}\int_{0}^{T} v(t)\,dt
\]

\paragraph{Maximum speed}
\[
v_{\max} = \max_{\,t\in[0,T]} v(t)
\]

\paragraph{Average acceleration}
\[
\bar{a} = \frac{1}{T}\int_{0}^{T} a(t)\,dt
\]

\paragraph{Maximum acceleration}
\[
a_{\max} = \max_{\,t\in[0,T]} a(t)
\]

\paragraph{Maximum change of direction}
This feature quantifies the attacker’s change of direction during the dribble. From the position coordinates, the player’s speed, acceleration, and jerk (the time derivative of acceleration) are computed. When the acceleration exceeds a threshold \(a_{\mathrm{thresh}} = 3 m/s^{2}\), it indicates the initiation of a directional turn. By analyzing the jerk \(j(t) = \frac{da}{dt}\), specifically the moments when acceleration increases (positive jerk) and then decreases (negative jerk), the start and end points of the turn, \(t_{\mathrm{start}}\) and \(t_{\mathrm{end}}\), are identified as:
\[
a(t) > a_{\mathrm{thresh}}, \quad
j(t_{\mathrm{start}}) > 0, \quad
j(t_{\mathrm{end}}) < 0.
\]
The movement direction at time \(t\) is defined by the angle
\[
\theta(t) = \arctan2\bigl(v_y(t), v_x(t)\bigr),
\]
where \(v_x(t) = \frac{dx}{dt}\) and \(v_y(t) = \frac{dy}{dt}\) are the velocity components. The degree of directional change is then quantified as the angular difference
\[
\Delta \theta = \left| \theta(t_{\mathrm{end}}) - \theta(t_{\mathrm{start}}) \right|.
\]
This approach follows the method introduced by Kai et al.~\cite{kai2021new}. Finally, the maximum change of direction is computed as 

\[
\theta_{\max} = \max_{\,t}\bigl|\theta(t+\delta t)-\theta(t)\bigr|.
\]

\subsection{2D Interactive Features}

The interactive features focus on the relationship between the attacker and the defender(s). According to Oonk et al.~\cite{oonk2024dribble}, these features also represent the group of variables that have the greatest impact on distinguishing successful from unsuccessful dribbles.

\paragraph{Relative speed} This feature describes the difference in speed between the attacker and the primary defender during the dribble. We consider both the average and maximal difference.
\[
\Delta v_{\max} = \max_{\,t\in[0,T]}\bigl[v_{\text{att}}(t)-v_{\text{def}}(t)\bigr]
\]

\[
\overline{\Delta v} = \frac{1}{T}\int_{0}^{T} \bigl[v_{\text{att}}(t)-v_{\text{def}}(t)\bigr]\,dt
\]

\paragraph{Pressure score}
The pressure exerted by nearby defenders on the attacker is determined using the pressure model of Andrienko et al.~\cite{andrienko2017visual}. Given this pressure score $P(t)$, we compute the average and maximum defensive pressure during the dribble as
\[
\bar{P}=\frac{1}{T}\int_{0}^{T} P(t)\,dt, 
\qquad 
P_{\max}= \max_{\,t\in[0,T]} P(t).
\]

\subsection{3D Ball Control}

\paragraph{Normalized ball-foot distance} 
Ball control describes how well a player keeps the ball within reach during the dribble. This feature is determined by measuring the average Euclidean distance between the ball and the foot that is closest to the ball.
\[
d_{\text{bf}} = \frac{1}{T}\int_{0}^{T} 
\bigl\|\mathbf{b}(t)-\mathbf{f}_{\min}(t)\bigr\|\,dt,
\]
where $\mathbf{b}(t)$ is the ball position and $\mathbf{f}_{\min}(t)$ is the toe joint closest to the ball.

How far a player can play the ball ahead of them is highly situation-dependent. When a player has a lot of space to dribble, they can play the ball further ahead without losing control of it. Therefore, we use the average pressure \( P \) exerted on the attacker during the dribble to normalize the ball control as follows:
\[
d_{\text{bf}}^{*} = d_{\text{bf}} \; P^{\alpha}, \qquad \alpha = 0.1.
\]

\paragraph{Ball touch frequency}
The frequency of ball contact was identified as a determinant of dribbling in sports science research~\cite{zago2016dribbling}. It also provides insight into the dribbling style employed: a higher frequency of ball contact generally points to a more technical dribbling style, whereas a lower frequency suggests players who tend to beat their opponent with speed.

The touch frequency can be derived from pose tracking data by analyzing the Euclidean distance between the ball and the big toe joint of the attacker’s nearest foot in each frame. A ball contact is detected when:
\begin{itemize}
  \item the distance is a local minimum, i.e., smaller than the distance in the preceding and following frame;
  \item the distance is smaller than a chosen threshold of 0.15 m;
  \item at least 0.25 seconds have passed since the previous detected ball contact, to avoid double-counting due to tracking noise.
\end{itemize}
The touch frequency is then calculated by dividing the number of ball contacts $N_{\text{touch}}$ by the total duration of the dribble $T$, yielding a value expressed in touches per second:
\[
f_{\text{touch}} = \frac{N_{\text{touch}}}{T}.
\]

\subsection{3D Balance Metrics}
Several studies have already shown that balance plays an important role in the technical skills of football players~\cite{evangelos2012proprioception,yusuf2022effect}. In addition, Zago et al.~\cite{zago2016dribbling} conclude that the centre of mass of players is also a determinant of dribbling

\paragraph{Centre-of-mass estimate}
Two approaches to determine a player’s centre of mass (CoM) based on the pose tracking data were implemented. A first, rather na\"ive, way to determine a player's centre of mass is simply to use the location of the \textit{midHip} joint. This joint is located just below the navel, which is generally considered representative of the CoM of the human body.

\[
\mathbf{CoM}_{\text{naive}} = \mathbf{j}_{\text{midHip}} .
\]

A second method estimates a player’s CoM using the anthropometric model of Winter~\cite{winter2009biomechanics}, which assigns each body segment a percentage of the total body weight.  In addition, it specifies where the CoM is located along each segment. Since these segments are defined by joints from the pose tracking data, their location is known, allowing us to estimate the CoM of a segment $i$ as:

\[
\mathbf{CoM}_i=(1-r_i)\,\mathbf{j}_{1,i}+r_i\,\mathbf{j}_{2,i},
\]

where $\mathbf{j}{1,i}$ and $\mathbf{j}{2,i}$ are the coordinates of the joints defining the segment and $r_i$ its CoM ratio. The total CoM of the player is then computed as:

\[
\mathbf{CoM}= \sum_{i=1}^{n} m_i\,\mathbf{CoM}_i,\qquad
\]

\paragraph{Imbalance}
This feature measures how centrally the player’s CoM is positioned above the base of support
and is defined as the distance in the XY-plane between the player's CoM and the midpoint between the two feet.
\[
I = \bigl\| \mathbf{CoM}_{x,y} - \mathbf{FMP}_{x,y} \bigr\|,
\]
where \(\mathbf{FMP}\) denotes the midpoint between the left and right foot joints.


\paragraph{Torso lean angle}
This feature indicates how much a player leans in a certain direction relative to their vertical axis, determined by defining a torso vector connecting the neck joint and the midpoint between the hips. The position of this vector in the vertical direction is then compared to its total length. The precise calculation is given by the formula:
\[
\theta_{\text{lean}}=
\arccos\!\left(\frac{z_{\text{neck}}-z_{\text{midHip}}}{\lVert
\mathbf{v}_{\text{torso}}\rVert}\right),\quad
\mathbf{v}_{\text{torso}}=\mathbf{j}_{\text{neck}}-\mathbf{j}_{\text{midHip}}.
\]
where \( \mathbf{j}_{\text{neck}} \) and \( \mathbf{j}_{\text{midHip}} \) denote the 3-D coordinates of the neck and midHip joints, respectively, and \( z_{\text{neck}} \) and \( z_{\text{midHip}} \) represent the vertical (z-axis) components of these joint coordinates

\subsection{3D Attacker–Defender Interaction}

\paragraph{Pass side vs. weighted leg}
At a given moment in a dribble, the attacker must choose to pass the defender on the left or right side. We hypothesize that it's easier to pass the defender on the opposite side of his weight-bearing leg. Therefore, we first identify the defender’s weight-bearing leg as based on the distances from his CoM to each heel joint. The leg with the shortest distance bears the most weight. Next, using 2D positional data, we infer the passing side by connecting the defender’s position to the center of their own goal and constructing a perpendicular boundary line. When the attacker crosses this line, the defender is considered passed. The attacker’s crossing relative to the defender’s left or right side determines the passing side. This leads to a binary feature

\[
\text{SidePass} =
\begin{cases}
1,&\text{attacker passes opposite the defender’s support leg},\\
0,&\text{otherwise}.
\end{cases}
\]

\paragraph{Defender stance angle}
We define a vector for both the attacker and the defender, starting at the left hip joint and ending at the right hip joint. By calculating the angle between these two vectors at the moment when the attacker passes the defender, we determine the defender’s posture as follows:
\[
\theta_{\text{defender-attacker}} = \arccos \left(
\frac{\mathbf{v}_{\text{defender}} \cdot \mathbf{v}_{\text{attacker}}}
{\|\mathbf{v}_{\text{defender}}\| \cdot \|\mathbf{v}_{\text{attacker}}\|}
\right)
\]
where $\mathbf{v}_{\text{defender}}$ and $\mathbf{v}_{\text{attacker}}$ are the hip vectors of the defender and attacker, respectively.

Based on this angle (ranging between 90° and 180°), three defender postures are distinguished:
\[
\left\{
\begin{array}{ll}
\text{side-on} & (90^{\circ} \!-\! 120^{\circ}) \\
\text{intermediate} & (120^{\circ} \!-\! 150^{\circ}) \\
\text{squared-up} & (150^{\circ} \!-\! 180^{\circ})
\end{array}
\right.
\]

%
\end{document}